\documentclass[11pt]{article}

\usepackage[preprint]{acl}
\usepackage{comment}
\usepackage{times}
\usepackage{latexsym}
\usepackage{multirow}
\usepackage[T1]{fontenc}

\usepackage[utf8]{inputenc}

\usepackage{microtype}

\usepackage{inconsolata}

\usepackage{graphicx}

\title{Language Modeling for the Future of Finance: A Survey into Metrics, Tasks, and Data Opportunities}

\author{Nikita Tatarinov \textsuperscript{$\spadesuit$ \Letter}, 
Siddhant Sukhani \textsuperscript{$\spadesuit$ \Letter}, 
Agam Shah \textsuperscript{\Letter},  
Sudheer Chava\\
Georgia Institute of Technology, Atlanta, USA \\
{\small  $\spadesuit$ \;Indicates equal contribution}\\
{\small \Letter \; Corresponding Authors: \{ntatarinov3, ssukhani3, ashah482\}@gatech.edu}
}

\usepackage{booktabs} 
\usepackage{graphicx} 
\usepackage{amssymb} 
\usepackage{marvosym} 
\usepackage{array} 

\definecolor{ForestGreen}{rgb}{0.13, 0.55, 0.13}
\newcommand{\gcheck}{\textcolor{ForestGreen}{\checkmark}} 

\begin{document}
\maketitle
\begin{abstract}
Recent advances in language modeling have led to a growing number of papers related to finance in top-tier Natural Language Processing (NLP) venues. To systematically examine this trend, we review 374 NLP research papers published between 2017 and 2024 across 38 conferences and workshops, with a focused analysis of 221 papers that directly address finance-related tasks. We evaluate these papers across 11 quantitative and qualitative dimensions, and our study identifies the following opportunities for NLP researchers: (i) expanding the scope of forecasting tasks; (ii) enriching evaluation with financial metrics; (iii) leveraging multilingual and crisis-period datasets; and (iv) balancing PLMs with efficient or interpretable alternatives. We identify actionable directions supported by dataset and tool recommendations, with implications for both the academia and industry communities.
\end{abstract}

\section{Introduction}

Language modeling is a core method in natural language processing (NLP) for analyzing unstructured text \cite{peters-etal-2018-deep, brown2020language}. At the same time, finance has become one of NLP’s primary application domains: as seen in Figure \ref{fig:code_available}, the number of \textit{finance-related papers in top-tier NLP venues} has been rising rapidly year over year. The tasks in these papers span general NLP problems on financial data, including sentiment analysis \citep{balakrishnan-etal-2022-exploring}, information extraction \citep{huang-etal-2023-fish}, and summarization \citep{khanna2022transformer}, as well as finance problems addressed with NLP techniques, such as stock prediction \citep{jain-agrawal-2024-fb}, volatility forecasting \citep{niu-etal-2023-kefvp}, etc. The primary goal of our study is to \textbf{highlight the opportunities to advance NLP research when applied to finance}.

Our scope includes 374 papers published from 2017 to 2024 across 38 NLP conferences and workshops. After further filtering (Section~\ref{sec:paper_selection}), we retain 221 papers that directly address finance-related tasks and evaluate them across 11 quantitative and qualitative dimensions, including tasks, methodologies, datasets, evaluation metrics, and accessibility. To the best of our knowledge, we present the first study to have a systematic look at finance-domain research in NLP venues, avoiding selection bias: existing surveys (Table~\ref{tab:lit_review}) manually select the reviewed papers. In addition, some works are focusing on specific NLP approaches such as deep learning \citep{OZBAYOGLU2020106384} and Large Language Models (LLMs) \citep{nie2024surveylargelanguagemodels,10.1145/3604237.3626869}, or on specific tasks such as sentiment analysis \citep{9142175}.

\begin{table*}[t]
\centering

\caption{Comparison of previous Natural Language Processing surveys in finance based on their focus areas, number of papers reviewed, and analytical features. Due to the absence of systematic collection methods in most prior research work, entries marked with (A) in the "Review Years" column indicate the approximate range.}

\resizebox{\textwidth}{!}{%
\begin{tabular}{%
  >{\centering\arraybackslash}m{0.25\textwidth}  
  >{\centering\arraybackslash}m{0.15\textwidth}  
  >{\centering\arraybackslash}m{0.12\textwidth}  
  >{\centering\arraybackslash}m{0.05\textwidth}  
  >{\centering\arraybackslash}m{0.25\textwidth}  
  >{\centering\arraybackslash}m{0.11\textwidth}  
  >{\centering\arraybackslash}m{0.07\textwidth}  
  >{\centering\arraybackslash}m{0.15\textwidth}  
  >{\centering\arraybackslash}m{0.10\textwidth}  
}
\toprule
\textbf{Paper} & \textbf{Review Years} & \textbf{Papers Reviewed} & \textbf{Features} & \textbf{Area surveyed} & \textbf{Systematic Collection} & \textbf{Domain trends} & \textbf{Quantitative Analysis} & \textbf{Temporal Analysis} \\
\midrule
\citep{10.1117/12.2604371} & 1959 (A)-2020 & 87 & 5 & General Overview & \texttimes & \texttimes & \texttimes & \texttimes \\
\citep{liu2024surveyfinancialaiarchitectures} & 2022-2024 & 49 & 11 & General Overview & \texttimes & \texttimes & \texttimes & \texttimes \\
\citep{millo2024integratingnaturallanguageprocessing} & 2018-2023 & 30 & 1 & Methodologies & \texttimes & \texttimes & \checkmark & \texttimes \\
\citep{chen2020nlpfintechapplicationspast} & 2016-2019 & 62 & 3 & Financial Technology & \texttimes & \texttimes & \texttimes & \texttimes \\
\citep{10.1145/3487553.3524868} & 2018-2022 & 38 & 2 & Financial Technology & \texttimes & \checkmark & \texttimes & \texttimes \\
\citep{Xing2017} & 1998 (A)-2016 & 127 & 4 & Financial Forecasting & \texttimes & \checkmark & \texttimes & \texttimes \\
\citep{zhao2024revolutionizingfinancellmsoverview} & 2004 (A)-2024 & 146 & 1 & Large Language Models & \texttimes & \texttimes & \checkmark & \texttimes \\
\citep{li2024largelanguagemodelsfinance} & 2020-2023 & 68 & 1 & Large Language Models & \texttimes & \texttimes & \texttimes & \texttimes \\
\citep{DONG2024100715} & 2023-2024 & 206 & 2 & Large Language Models & \texttimes & \texttimes & \checkmark & \checkmark \\
\citep{nie2024surveylargelanguagemodels} & 2019-2024 & 318 & 1 & Large Language Models & \texttimes & \texttimes & \texttimes & \checkmark \\
\citep{Lee_2025} & 2018-2023 & 51 & 3 & Large Language Models & \texttimes & \texttimes & \texttimes & \texttimes \\
\citep{8780312} & 2004-2019 & 89 & 1 & Machine Learning & \texttimes & \texttimes & \texttimes & \texttimes \\
\citep{OZBAYOGLU2020106384} & 1998 (A)-2020 & 151 & 6 & Deep Learning & \texttimes & \texttimes & \checkmark & \checkmark \\
\citep{9142175} & 2003 (A)-2020 & 89 & 1 & Sentiment Analysis & \texttimes & \checkmark & \texttimes & \checkmark \\
\midrule
\textbf{Language Modeling for the Future of Finance} & \textbf{2017-2024} & \textbf{374} & \textbf{11} & \textbf{Tasks, Methodologies, Data, Metrics, Code, Authorship, Funding} & \gcheck & \gcheck  & \gcheck  & \gcheck \\
\bottomrule
\end{tabular}%
}
\label{tab:lit_review}

\end{table*}

Our analysis reveals not only valuable insights but also actionable directions for both research and practice with dataset and tool recommendations. Financial forecasting tasks, while well-established, leave room for exploration, especially in areas such as risk and macroeconomic prediction (Section \ref{sec:task_distribution}). Financial evaluation metrics are gaining traction and could further improve the practical applicability of models (Section \ref{subsec:metrics}). The growing availability of temporally diverse (Sections \ref{subsec:crisis_years}, \ref{subsec:data_sources}), multilingual and multimodal (Sections \ref{subsec:data_biases}) datasets enables more robust, globally applicable models.

\section{Paper Extraction Process}
\label{sec:paper_selection}

\begin{figure*}[ht]
    \centering
    \includegraphics[width=\linewidth]{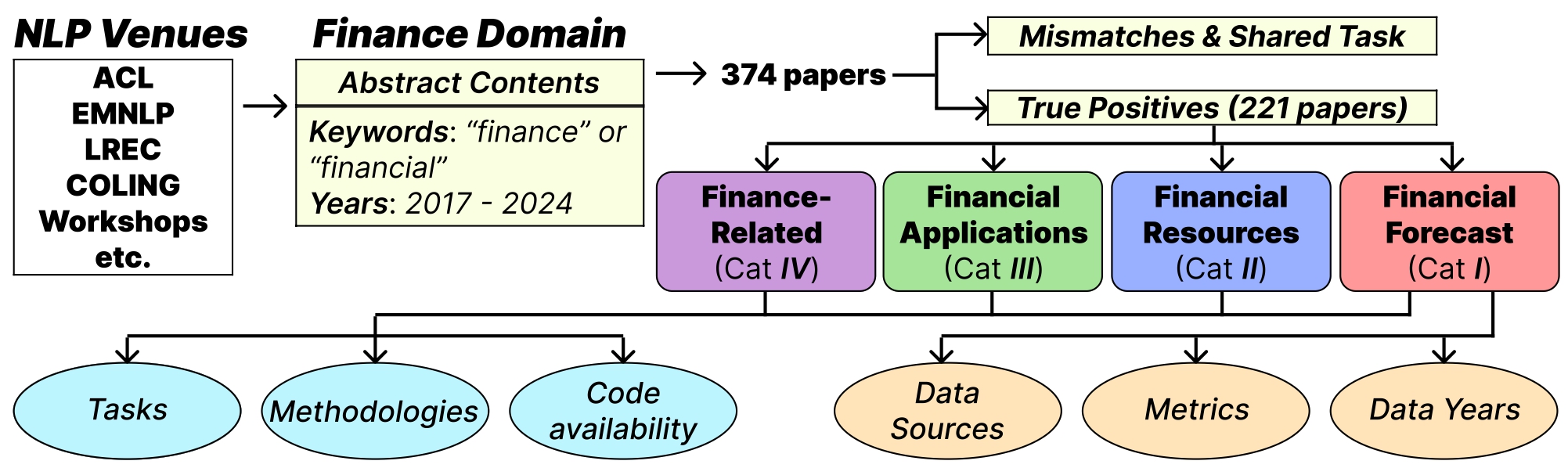}
    \caption{Overview of our paper selection process and analysis dimensions. We collected papers from a broad range of NLP venues using abstract-level keyword filtering, yielding 374 candidates. After removing mismatches and shared task papers, we retained 221 papers, categorized into four groups by their connection to financial tasks.}
    \label{fig:paper_selection}
\end{figure*}

To study how NLP has been applied to finance, we selected papers from 38 NLP venues, including ACL, NAACL, EMNLP, LREC, CoLM, COLING, and workshops like FinNLP and the Workshop on Economics and Natural Language Processing. As illustrated in Figure \ref{fig:paper_selection}, we began by filtering papers that mentioned “finance” or “financial” in their abstracts \cite{Mackenzie2018QueryEarlyStage, Nogueira2020MonoT5}, enabling us to cast a wide net, allowing for a high volume of papers that could be manually postprocessed. Out of all such papers published from 1975 to the present, more than 94\% appeared from 2017 onward, coinciding with the emergence of transformer-based models \citep{vaswani2023attentionneed}, which greatly expanded the scope of NLP applications \cite{devlin-etal-2019-bert, dai-etal-2019-transformer, lewis-etal-2020-bart}, and we therefore use 2017 as a threshold year for our analysis.

This initial filtering returned 374 papers. Among them, 88 were tied to shared tasks (e.g., SemEval-2017 \citep{kar-etal-2017-ritual}, FinCausal-2022 \citep{mondal-etal-2022-expertneurons}), and 65 used the term “financial” in a different contexts (such as referencing financial resources \citep{sekeres-etal-2024-developing,ding-riloff-2018-human}) without any actual financial application. After removing these, we retained 221 papers for our analysis.

Unlike previous surveys as shown in Table \ref{tab:lit_review}, which often emphasize qualitative observations \cite{10.1117/12.2604371, liu2024surveyfinancialaiarchitectures, millo2024integratingnaturallanguageprocessing}, our study combines quantitative and qualitative methods for a broader view of the field. In comparison to other works, we categorized the selected papers by their primary focus into four task-based categories (Figure \ref{fig:task_distribution_by_level}). This classification helps us examine trends in how NLP techniques are used to tackle finance-relevant problems and supports more detailed comparisons across types of tasks.

\section{Task Distribution in NLP for Financial Applications}
\label{sec:task_distribution}

\begin{figure*}[t]
    \centering
    \includegraphics[width = 0.95\textwidth]{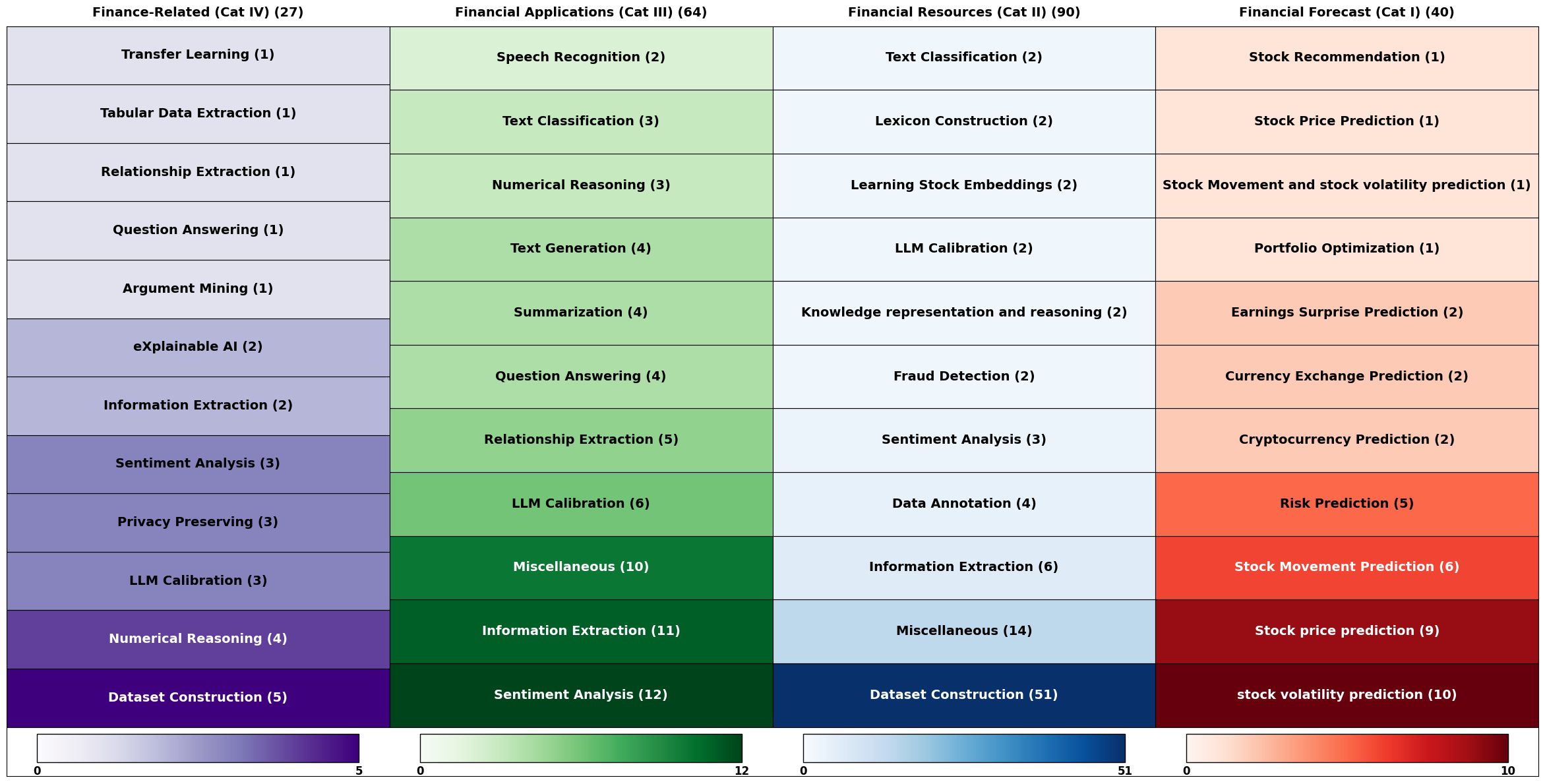}
    \caption{Distribution of primary tasks across categories. Each cell shows the task name and paper count (e.g., "Sentiment Analysis (3)"), with color gradients indicating frequency -- darker shades represent more papers. "Miscellaneous" groups tasks that appear only once within Categories II and III.}
    \label{fig:task_distribution_by_level}
\end{figure*}

To understand how NLP has been used in financial contexts, we categorized papers into four distinct groups (shown in Table \ref{tab:categories}) based on their primary tasks (Figure \ref{fig:task_distribution_by_level}).

\begin{table}[htbp]
\centering

\caption{Summary of task categories based on the specifics of their financial focus, ranging from direct prediction of financial outcomes to general NLP tasks with potential financial value.}
\label{tab:categories}

\resizebox{\linewidth}{!}{
\begin{tabular}{p{0.3\linewidth} p{0.65\linewidth}}
\hline
\textbf{Category} & \textbf{Description} \\
\hline
\textcolor{red}{\textbf{Financial Forecast (Category I)}} & Targets predicting financial events, including stock movements, volatility, bankruptcy, and currency exchange rates. \\
\textcolor{blue}{\textbf{Financial Resources (Category II)}} & Covers tasks addressing finance-specific issues beyond prediction, such as constructing financial datasets, detecting fraud in finance-related documents, and extracting financial events. \\
\textcolor[RGB]{50,255,50}{\textbf{Financial Applications (Category III)}} & Focuses on general ML/NLP tasks like sentiment analysis and information/relation extraction for financial datasets. \\
\textcolor[RGB]{150,25,255}{\textbf{Finance Related (Category IV)}} & Covers tasks not applied to financial data or targeting financial problems, but potentially useful in finance, such privacy-preserving and explainable AI. \\
\hline
\end{tabular}
}

\end{table}

\textcolor{red}{\textbf{Financial Forecast (Category I)}} papers cover predictive tasks such as stock price and volatility forecasting. While these tasks are well-studied, areas like economic forecasting \citep{arno-etal-2023-numbers}, risk assessment \citep{zhou-etal-2020-interpretable}, and cryptocurrency prediction \citep{seroyizhko-etal-2022-sentiment} receive less focus. These areas offer opportunities to expand the reach of predictive models, improving robustness and comparability (Section \ref{sec:concerns}).

\textcolor{blue}{\textbf{Financial Resources (Category II)}} papers often focus on dataset construction (Figure \ref{fig:timeline}). Ranging from annotated earnings calls to news and speeches, These datasets support tasks like financial event extraction \citep{huang-etal-2024-extracting,ju-etal-2023-compare}, fraud detection \citep{erben-waldis-2024-scamspot,wang-etal-2019-real}, and annotation \citep{aguda-etal-2024-large,khatuya-etal-2024-parameter}. While these tasks enhance financial data processing, other tasks remain relatively underrepresented.

In the \textcolor[RGB]{50,255,50}{\textbf{Financial Applications (Category III)}} group, sentiment analysis \citep{rodriguez-inserte-etal-2023-large}, information extraction \citep{lior-etal-2024-leveraging}, and question answering \citep{kosireddy-etal-2024-exploring} are the dominant tasks. These methods are widely used to analyze earnings calls, reports, and market commentary, where investor sentiment and factual extraction are key inputs for decision-making\cite{chen-etal-2021-finqa, zhu-etal-2021-tat, mukherjee-etal-2022-ectsum,qin-yang-2019-say}. Recent efforts have aimed to improve QA models for handling financial text \citep{theuma-shareghi-2024-equipping,liu-etal-2024-beyond,mavi-etal-2023-retrieval}.

Finally, among \textcolor[RGB]{150,25,255}{\textbf{Finance-Related (Category IV)}} papers, the most studied areas are dataset construction and numerical reasoning. While these datasets are not finance-specific, they have potential applications in finance. For instance, fake news detection datasets \citep{vargas-etal-2021-toward-discourse} can help reduce misinformation in markets. Numerical reasoning \citep{akhtar-etal-2023-exploring}, important for understanding financial statements and assessing risk, is receiving more attention. However, other areas -- such as explainable AI (XAI) \citep{KLEIN2024106358} and privacy-preserving methods \citep{10.1257/aer.102.3.65} remain rarely explored, despite their relevance to secure and interpretable decision-making \citep{basu-etal-2021-privacy}.

\section{Potential in Financial Forecasting}
\label{sec:concerns}

\textcolor{red}{\textbf{Category I}} papers reveal several areas where forecasting models could be improved. Most studies use general ML metrics, leaving out finance-specific measures that better capture practical utility (Section ~\ref{subsec:metrics}). Crisis periods are rarely considered, limiting insights into how models behave under stress (Section ~\ref{subsec:crisis_years}). Finally, there is a strong U.S. and English-language bias in data, with limited adoption of global financial datasets (Section ~\ref{subsec:data_sources}, ~\ref{subsec:data_biases}). Expanding these areas of exploration could strengthen model robustness and support broader real-world applicability.

\subsection{Advancing Evaluation with Finance Specific Metrics}
\label{subsec:metrics}

\begin{figure}[t]
    \centering
    \includegraphics[width=\linewidth]{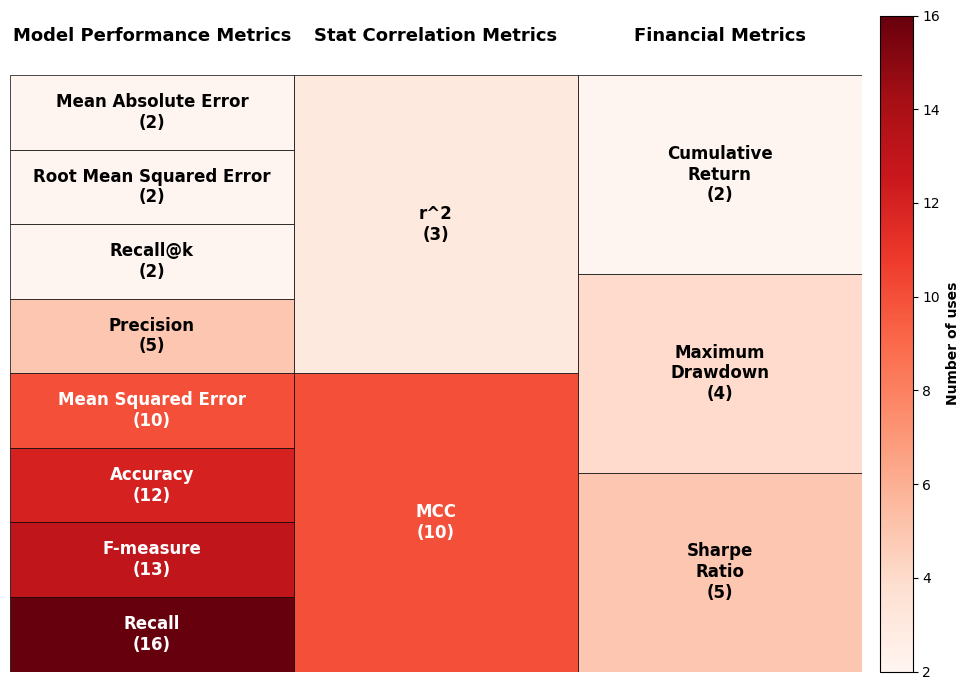}
    \caption{Distribution of evaluation metrics used in \textcolor{red}{\textbf{Category I}} papers. Most rely on ML-based metrics, while only a few financial metrics appear repeatedly.}
    \label{fig:metrics}
\end{figure}

As shown in Figure \ref{fig:metrics}, \textcolor{red}{\textbf{Category I}} papers mostly rely on ML metrics such as accuracy, F1, or MSE \citep{sawhney-etal-2020-voltage,wu-2020-event,yangjia-etal-2022-fundamental}. While useful, these do not fully reflect financial performance. Finance-specific metrics offer risk-adjusted insights that ML measures overlook. Incorporating such metrics would improve \textit{comparability across models and align research outcomes with practical needs} \citep{zhang-etal-2024-finbpm, zou-etal-2022-astock,sawhney-etal-2021-quantitative}.

As highlighted by \citet{Tamar2012VarianceRiskPG, Liu2022FinRLMETA}, the financial machine learning literature stresses that conventional statistical metrics such as mean squared error are insufficient for assessing trading strategies, since they overlook the dimensions of real-world profitability and risk management \citep{bailey2015probability, deprado2018advances}. Instead, financial metrics such as Sharpe Ratio, Maximum Drawdown, and Cumulative Return are essential because they reflect the model’s risk-adjusted performance and robustness across market regimes \citep{lo2002statistics, krauss2017deep, takahashi2009designing} as well as the applicability of the model to practical real world situations.

\paragraph{Key Financial Metrics and Python Libraries}
In financial machine learning, specialized metrics are essential for evaluating a model’s real-world applicability. The \textit{Sharpe Ratio} quantifies risk-adjusted return by comparing excess returns to volatility \citep{Investopedia_SharpeRatio}, while \textit{Maximum Drawdown (MDD)} captures the largest peak-to-trough loss, reflecting downside risk and robustness \citep{investopediaMaxDrawdown}. Similarly, \textit{Cumulative Return} provides a direct benchmark by measuring total profit over a period \citep{investopediaCumulativeReturn}. Several Python libraries facilitate the computation of these metrics within model-driven investment workflows: \texttt{PyPortfolioOpt} \citep{Martin2021} offers portfolio optimization tools including Sharpe and risk-return analysis; \texttt{QuantLib} \citep{quantlib} supports pricing, drawdown, and advanced risk modeling; \texttt{Backtrader} \citep{backtrader} enables backtesting and performance evaluation of trading strategies, including NLP-driven pipelines; and \texttt{bt} \citep{morissette_bt} streamlines prototyping and comparison of forecasting models with built-in financial metrics.

\subsection{Strengthening Robustness by Incorporating Crisis Periods}
\label{subsec:crisis_years}

\begin{figure*}[t]
    \centering
    \includegraphics[width=0.95\linewidth]{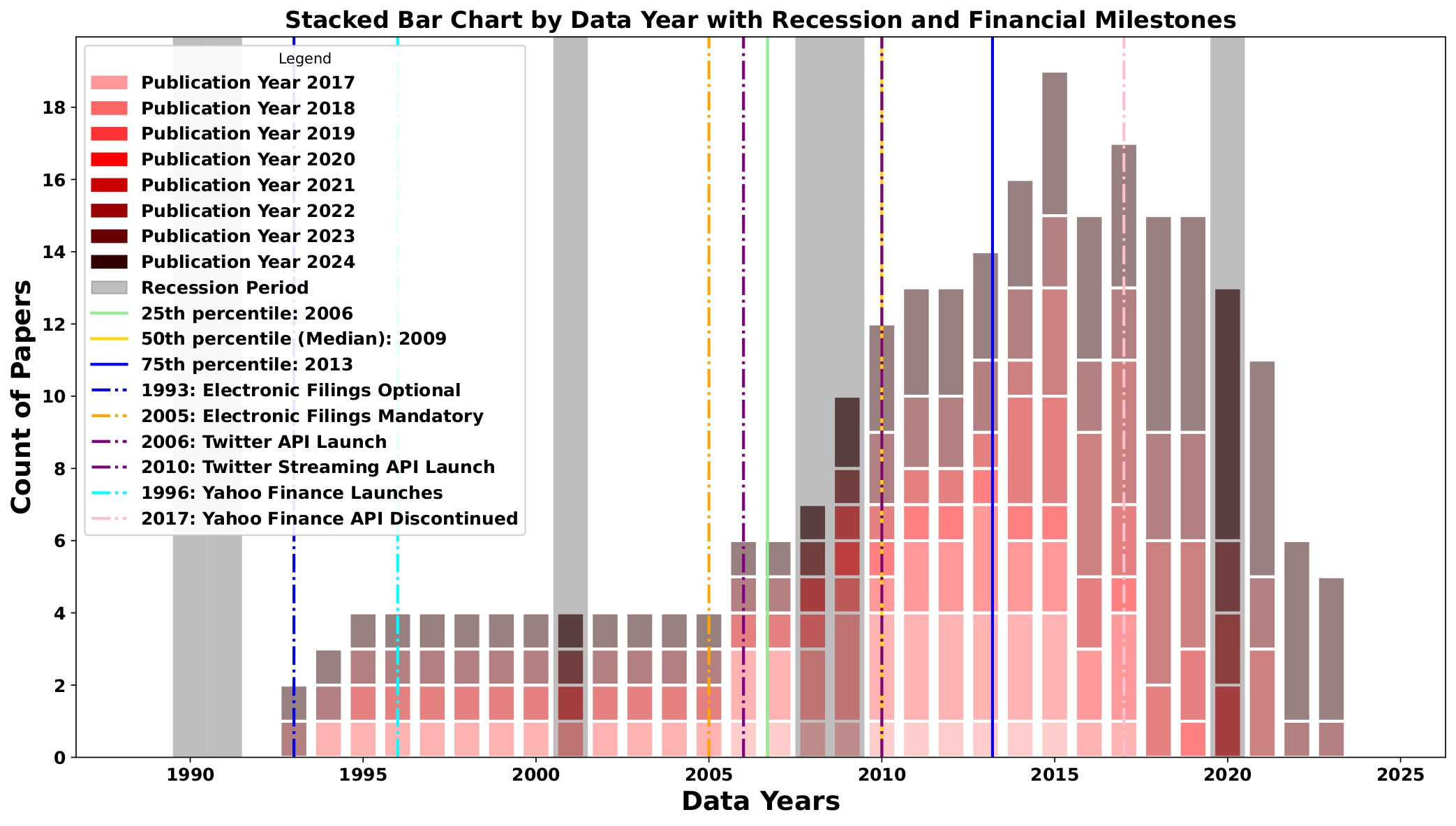}
    \caption{Data year distribution in financial forecasting papers, annotated with major financial events and infrastructure milestones. Highlights underuse of crisis periods despite their importance for model robustness.}
    \label{fig:kde years}
\end{figure*}

Figure \ref{fig:kde years} shows that researchers use financial data dating back to 1993, when electronic filings became publicly available, with another rise in 2005 after the HTML filing mandate \citep{SEC_EDGAR_Overview}. However, most publications use post-2009 (the global financial crisis) and pre-2020 (the COVID-19 pandemic) data, thus missing the crisis periods in evaluation \cite{Kim2022RevIN, Fan2023DishTS, Yao2022WildTime}. \textit{Crisis periods are essential for evaluating model robustness, allowing stress testing} \citep{Investopedia_StressTesting, Hafiz2023NeuralCrisis, Aldasoro2025MarketStressML} and helping build forecasting models that remain reliable under instability.

For example within industrial applications of such models, LTCM’s overdependence on stable market data and highly leveraged VaR models left it blind to tail risks, so when the Asian and Russian crises hit, spreads exploded, liquidity vanished, and the fund lost over 90\% of its equity, sparking systemic risk concerns \citep{cambridgeLTCM2003,frbLTCMHistory,vanityfair1998LTCM}. While there were data availability challenges prior to 2015 \citep{Bloomberg_DataForGood_2015}, the crisis years remain underused. In Section \ref{subsec:data_sources}, we highlight valuable data sources, including those with crisis years data.

\subsection{Opportunities in Data Coverage}
\label{subsec:data_sources}

Most forecasting models use common financial sources: stock prices, SEC filings, financial news, earnings calls. But many valuable datasets are either rarely used (e.g., Federal Reserve reports \citep{shah-etal-2023-trillion, menzio-etal-2024-unveiling}) or completely absent from NLP research (e.g., shareholder letters). Public resources like FRED \citep{FRED} and Fama-French \citep{FamaFrench} offer rich but underexplored macro indicators. Incorporating these datasets into ablation studies and further analysis could help align the various findings with macroeconomic trends and market dynamics, improving model reliability \cite{chakraborty2016predicting, xu2018stock,sawhney-etal-2020-voltage}. We highlight sources of valuable financial data, including those already used by the community and the other beneficial ones in Appendix \ref{app:DataSources}.

\subsection{Data Concerns and Biases}
\label{subsec:data_biases}
\paragraph{Data Accessibility Concerns}
Despite the presence of a large variety of data sources, there are significant limitations in data accessibility. Legacy datasets described in Section \ref{subsec:data_sources} often become outdated, and APIs for accessing data may be discontinued, as the Yahoo Finance API \citep{yahooyfinance_discontinued}, or restricted by paywalls. Other essential databases, such as CRSP, require paid access and are unavailable to most individual researchers. These barriers severely limit reproducibility and access to reliable data, underscoring the need for open, regularly updated financial datasets for forecasting.

\paragraph{Data Language and Country Bias}

As shown in the previous sections, financial NLP research is heavily skewed toward English-language data, particularly from U.S. markets \citep{chen-etal-2021-finqa, reddy-etal-2024-docfinqa, mukherjee-etal-2022-ectsum}. Most forecasting datasets rely on U.S. regulatory filings, earnings calls, and financial news, which is often centered on large-cap firms like those in the S\&P 500. This narrow focus limits the applicability of models to non-English and non-U.S. contexts. Figure \ref{fig:timeline} highlights that while some multilingual datasets exist, they are mostly designed for language modeling rather than financial forecasting \cite{ding-etal-2014-using, cortis-etal-2017-semeval}. Very few datasets enable forecasting tasks in other languages, with rare exceptions like the bilingual CMIN corpus \citep{luo-etal-2023-causality}. \textit{Expanding multilingual, market-diverse datasets is essential for building globally robust forecasting models} and mitigating systemic biases in NLP for finance, as also noted by \citep{jorgensen-etal-2023-multifin}.

\paragraph{Multimodal Data Scarcity}
Despite significant progress in multimodal learning, the domain of NLP in Finance still lacks comprehensive datasets that integrate text, audio, and other critical financial modalities \cite{sawhney-etal-2020-voltage, sawhney-etal-2021-multimodal, Kaikaus2022TruthOrFiction}. MAEC, a dataset aligning transcripts with audio from over 3,400 S\&P 500 earnings calls (920+ hours of speech) for financial risk forecasting, and FinAudio, the first standardized benchmark for audio-first tasks with 400+ hours of financial audio for evaluating Audio-LLM capabilities, represent two of the key recent resources in financial audio research \citep{Li2020MAEC,cao2025finaudiobenchmarkaudiolarge}. Our survey indicates that while many existing datasets support exploration of acoustic and textual cues in financial forecasting, truly multimodal resources (those integrating visual data such as stock charts, presentation slides alongside structured financial metrics) remain very limited. Notable exceptions include \citet{10.1145/3503161.3548380} and \citet{fons-etal-2024-evaluating}, which begin to address this gap. This modality gap highlights the pressing need for richer, tri- or quad-modal datasets to better capture the multifaceted signals that influence market dynamics \cite{Liang2022ModalityGap, Li2024TimeMMD}.

\section{Broader Insights for NLP in Finance}
\label{sec:broader_insights}

\subsection{Complementing Foundation Models with Practical Alternatives}
\label{subsec:plm_limitations}

Based on our observations from our vast corpus of research, the methodological landscape in NLP for Finance has shifted decisively toward pretrained language models (PLMs) and large language models (LLMs). We group existing approaches into several broad categories, which allows us to track adoption trends and the rise of PLMs, beginning with the use of BERT-based models in finance around 2019 \citep{araci2019finbert, yang2020finbert}, and accelerating rapidly with the emergence of general-purpose LLMs like ChatGPT \citep{wu2023chatgptfinance, gao2023llmtrends}. However, NLP tasks in Finance often pose domain-specific constraints such as interpretability, regulatory transparency, data scarcity, and the need for low-latency systems \cite{Luo2018InterpretableFinance, Maia2018FiQA}. These factors make non-PLM methods (such as rule-based classifiers, feature-engineered models, and domain-specific embeddings) not only viable but often preferable in production environments \citep{shah-etal-2023-trillion, lopez2021nonnlpfinance}. The rest of this section revisits key methodological alternatives to PLMs, each addressing practical and conceptual needs that remain critical in industrial financial tasks.

\paragraph{Latency and Scalability}
Classical NLP pipelines are far more resource-efficient than LLMs \cite{strubell-etal-2019-energy}. In time-time contexts like news analytics or alert systems, lightweight models (e.g., logistic regression or small CNNs) deliver insights instantly and can scale to massive corpora without requiring GPU infrastructure \citep{zhai2019charting}. Systems like RavenPack \citep{ravenpack} and Refinitiv News Analytics (TRNA) \citep{trna} use rule-based NLP and horizontally scalable architectures to deliver structured news data with sub-second latency, supporting real-time trading and high-throughput analysis.

\paragraph{Interpretability and Compliance}
Lexicon- and rule-based methods remain essential in regulated settings \cite{moreno-ortiz-etal-2020-design, du2023finsenticnet}. Financial sentiment lexicons (e.g., Loughran-McDonald) \citep{loughran2011liability} and keyword rules are transparent, auditable, and often outperform neural models in compliance or risk-flagging tasks \citep{hosseini2018auditnlp}. Lexicon-based surveillance tools, such as those used in communications monitoring \citep{communications_surveillance} and financial crime compliance \citep{lexicon_surveillance}, are widely adopted for their transparency and ease of audit, allowing firms to justify alerts with specific keywords.

\paragraph{Information Extraction and Knowledge Graphs}
Named-entity and relation extraction systems power knowledge graphs and schema-driven applications \cite{elhammadi-etal-2020-high, hamad-etal-2024-fire}. These systems offer precision, structure, and explainability, which are ideal for fraud detection, Know Your Customer (KYC), and regulatory alignment, where PLMs often lack controllability \citep{szarvas2007unsupervised, tarnopolski2019knowledgegraphs}. Financial institutions use rule-based NLP for entity and relation extraction to build and update knowledge graphs in KYC and Anti-Money Laundering (AML) workflows, as seen in solutions from \citet{kychub} and \citet{tigergraph}, enabling automated risk monitoring and explainable link discovery.

\paragraph{Data Efficiency and Domain Adaptability}
Feature-based pipelines using TF-IDF, static embeddings, and classical classifiers (SVMs, XGBoost) work well with limited data\cite{wang-manning-2012-baselines}. They enable explicit domain knowledge integration (such as tagging financial terms or numerical cues) for credit scoring, bankruptcy prediction, and event detection \citep{li2018credit, khandani2021practical, alanis2022benchmarking}. Financial firms often use TF‑IDF \citep{capitalone_tfidf} and XGBoost \citep{xgboost_case_study} to build effective models with limited labeled data without requiring costly infrastructure.

\paragraph{Structure and Summarization}
Unsupervised techniques like LDA and extractive summarization remain valuable for discovering latent topics or summarizing dense financial documents \cite{filippova-etal-2009-company, agrawal2021goal}. These methods are fast, interpretable, and competitive on factual corpora like earnings calls or filings \citep{blei2003latent, nenkova2011automatic}. Companies like Bank of America \citep{lda_on_practice} and American Express \citep{arora-radhakrishnan-2020-amex} apply topic modeling and extractive summarization to earnings calls and filings, using unsupervised or minimally supervised methods for reliable and traceable outputs.

\subsection{Implications of Shift Toward General-Purpose Models}
\label{subsec:plm_shift_implications}

General-purpose language models have rapidly become the default choice in NLP applied to Finance, with models like RoBERTa \citep{liu2019robertarobustlyoptimizedbert}, GPT-4 \citep{achiam2023gpt}, and LLaMA-2 \citep{touvron2023llama} adopted far more widely and quickly than domain-specific alternatives. As Figure \ref{fig:timeline} shows, even earlier versions of FinBERT \citep{araci2019finbertfinancialsentimentanalysis,yang-etal-2020-finbert,liu-etal-2021-finbert,huang-etal-2022-finbert} are used more than newer ones, and recent finance-tuned models like FinGPT \citep{yang-etal-2024-fingpt} remain rare. For practitioners, this means \textit{many models are optimized for benchmarks, not for the realities of finance}, where interpretability, latency, or regulation often matter more than raw accuracy.

\begin{figure}[t]
    \centering
    \includegraphics[width=\linewidth]{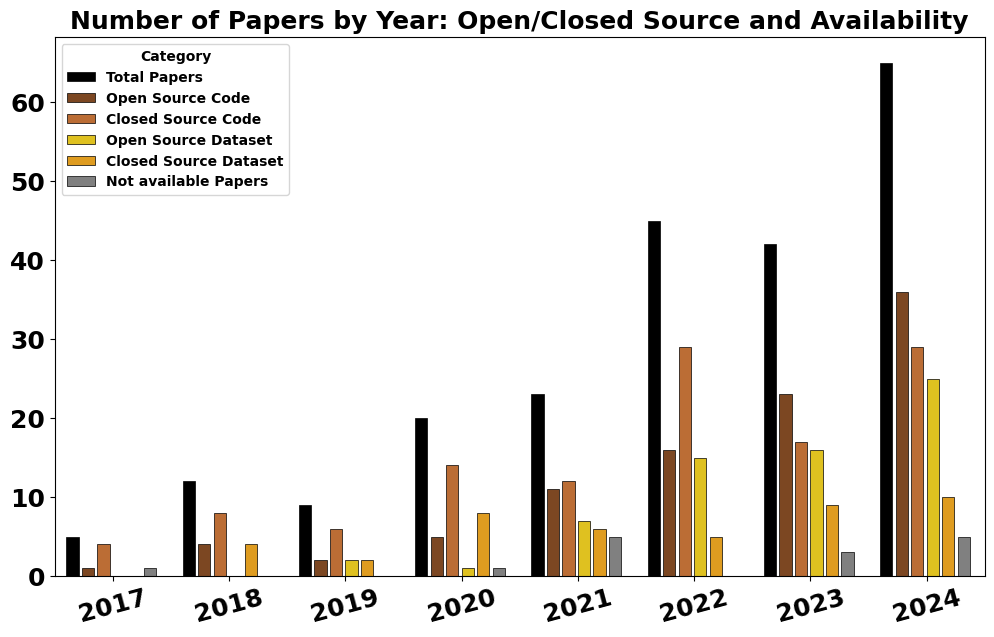}
    \label{sec:code_availability}
    \caption{Trends in code and dataset availability, highlighting the shift toward open-source practices and the growing accessibility of NLP resources for finance.}
    \label{fig:code_available}
\end{figure}

\begin{table*}[htbp]
\centering
\tiny
\caption{Summary of trends and research directions in NLP research applied to finance, highlighting areas of focus, methodological shifts, and key opportunities for future exploration.}
\label{tab:financial_nlp_summary}
\footnotesize
\begin{tabular}{%
  >{\arraybackslash}p{0.1\textwidth}  
  >{\arraybackslash}p{0.4\textwidth}  
  >{\arraybackslash}p{0.4\textwidth}
}
\toprule
\textbf{Criteria} & \textbf{Trends and Observations} & \textbf{Potential Opportunities and Recommendations} \\
\midrule
\textbf{NLP Tasks in Finance} & Sentiment analysis, information extraction, and question answering are the most frequently addressed tasks. Forecasting is mostly centered around stock and volatility prediction. & Underexplored areas such as explainability, privacy-preserving methods, and tasks like bankruptcy or cryptocurrency prediction present valuable directions for future work (Section \ref{sec:task_distribution}). \\

\textbf{Evaluation Metrics} & Most studies rely on ML metrics such as accuracy and MSE, which do not reflect financial performance. &  Incorporating financial metrics like Sharpe Ratio or Maximum Drawdown would improve the practical relevance and comparability of predictive models (Section \ref{subsec:metrics}). \\

\textbf{Crisis Periods} & Few studies include data from financial crises such as 2008–2009 or 2020–2021, focusing instead on stable periods. & Including data from volatile periods can support stress-testing and help build models more resilient to real-world fluctuations (Section \ref{subsec:crisis_years}). \\

\textbf{Data Selection and Bias} & Most studies use English-language datasets and U.S. financial sources. While there is growing diversity in the types of data used (e.g., news, filings, social media), much of it is static, with limited updates or adaptation to changing financial contexts. & Developing multilingual and globally representative datasets (Section \ref{subsec:data_biases}), along with maintaining and updating existing ones (Section \ref{subsec:data_sources}), would support better generalization and long-term applicability of models across financial domains. \\

\textbf{PLM/LLM Adoption} & General-purpose PLMs and LLMs have largely replaced finance-specific and custom architectures. At the same time, the use of statistical NLP and conventional ML continues to decline. & Revisiting finance-specific models and exploring alternative methods such as graph-based learning or hybrid statistical models could offer improvements for financial tasks (Sections \ref{subsec:plm_limitations}, \ref{subsec:plm_shift_implications}). \\

\textbf{Open Accessibility} & Code and dataset sharing has become more common, especially after 2021, enhancing transparency and reproducibility. & Open access should be a standard in NLP-for-finance research when legally feasible. Maintaining functional and up-to-date repositories is key for long-term reproducibility (Section \ref{subsec:code_data_accessibility}). \\

\bottomrule
\end{tabular}%
\end{table*}

This shift is also mirrored in the decline of custom architectures. Prior to 2022, task-specific models were more common: TagOp for TAT-QA \citep{zhu-etal-2021-tat}, HyBrider for HybridQA \citep{chen-etal-2020-hybridqa}); MDRM \citep{qin-yang-2019-say}, HTML \citep{yang-etal-2020-html}, and HAN \citep{hu-etal-2021-han} in volatility prediction \citep{niu-etal-2023-kefvp,mathur-etal-2022-docfin}). Today, these efforts are rarely pursued. However, as Section \ref{subsec:plm_limitations} discusses, real-world financial applications often require design trade-offs, such as transparency for compliance, or fast inference for deployment, and off-the-shelf LLMs do not accommodate them well. \textit{The move away from custom models narrows the solution space} at a time when financial tasks still demand methodological flexibility.

\subsection{Enabling Reproducibility through Open Resources}
\label{subsec:code_data_accessibility}

As we witness an accelerated adoption of language models, the issue of open accessibility in NLP research, especially in finance, becomes more relevant. As shown in Figure \ref{fig:code_available}, open-source practices have become more common in recent years. Before 2021, closed-source code dominated, but the trend has shifted toward sharing code and datasets.

This move enhances transparency, reproducibility, and collaborative progress in NLP research applied to finance \citep{Whited2023Costs}. Despite this, some papers still include inactive links or empty repositories, limiting reproducibility. At the same time, dataset availability has steadily increased since 2017, pointing to growing awareness around open data. \textit{Maintaining functional, up-to-date repositories remains essential to support meaningful benchmarking and model development}.

\section{Conclusion \& Discussion}
\label{sec:conclussion}

Drawing on Table \ref{tab:financial_nlp_summary}, our survey shows that sentiment analysis, information extraction, and forecasting have driven NLP in finance, yet critical gaps remain: finance‑specific evaluation metrics, stress‑testing on crisis data, and truly global, multilingual datasets. These findings underscore both the field’s progress and the work ahead to build resilient, transparent, and inclusive financial NLP solutions.

\paragraph{Implications for NLP and Language Modeling Researchers} The application of LMs to financial domains presents challenges that differ from traditional NLP benchmarks. Our findings suggest that domain-specific requirements, such as interpretability, latency, and compliance, make classical or hybrid approaches highly relevant. The lack of financial evaluation metrics, underuse of crisis-period data, and limited attention to multilingual and global financial corpora signal areas where NLP researchers can contribute meaningfully. Moreover, the scarcity of domain-adapted models points to missed opportunities for more effective domain transfer. As a result, our observations indicate that future work should prioritize robustness, transparency, and real-world relevance over benchmark performance alone.

\paragraph{Considerations for Financial Practitioners} For finance professionals and institutions considering NLP solutions, this survey provides guidance on selecting appropriate tools. Off-the-shelf LLMs may appear attractive, but their performance in finance-specific contexts is often limited by lack of customization, interpretability, or latency guarantees. Classical and lexicon-based methods, when tailored to regulatory or operational constraints, can outperform black-box models in compliance or auditing settings. Practitioners should also evaluate model performance using finance-aligned metrics and test under historical stress scenarios. Finally, leveraging multilingual datasets and broadening market coverage can help mitigate geographic and systemic biases in automated decision-making.

\section*{Limitations}
\label{sec:limitations}

For readability, we standardize task names and merge closely related variants when counting. When a paper spans multiple areas, we assign a single primary tag to keep statistics interpretable; a multi-tag alternative would yield slightly different totals. These are intentional choices to aid comparability.

\section*{Ethical Considerations}
\label{sec:ethics}

This study does not assess the merit or quality of individual papers. We do not suggest that any category or method is inherently better than others. Our goal is to map the volume and distribution of research applying NLP methods to finance, and to offer a foundation for further study, without making judgments about the value of specific approaches.

\bibliography{anthology,references}
\clearpage
\appendix

\section{Timeline Figure}
\label{app:timeline_fig}

\begin{figure*}[htbp]
    \centering
    \includegraphics[width=\textwidth]{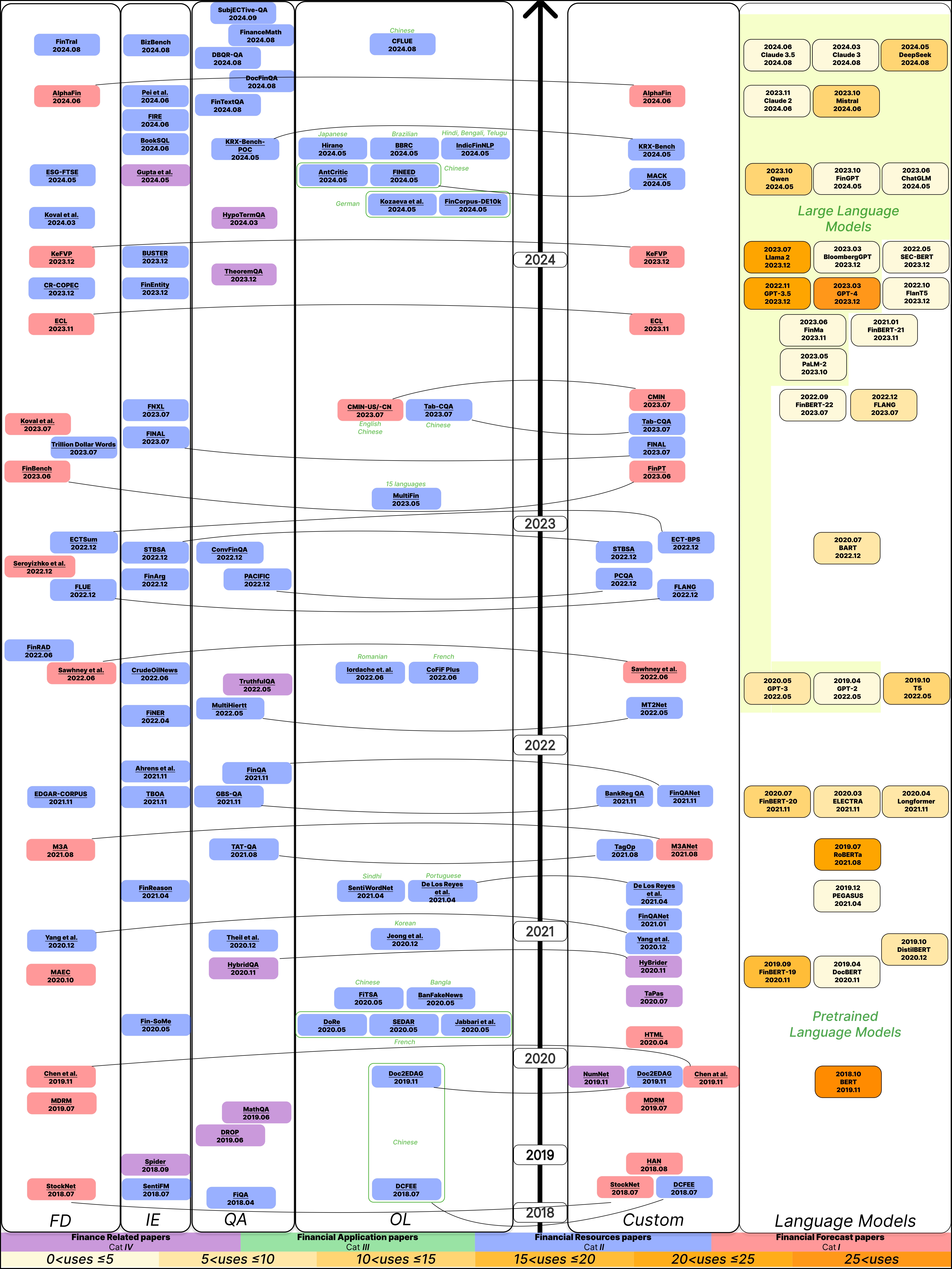}
    \caption{Timeline of PLM/LLM adoption in NLP research applied to finance, alongside key datasets by task type: C (custom models), QA (question answering), IE (information extraction), OL (other language), and FD (financial documents). For each model, the top date indicates release; the bottom, first usage in the surveyed papers.}
    \label{fig:timeline}
\end{figure*}

This appendix includes the timeline Figure \ref{fig:timeline}.

\section{Authorship and Funding}
\label{app:authorship_funding}

Based on the analysis we conducted, we also made a binary classification of the following categories:

\begin{itemize}
    \item Academic authors
    \item Industrial authors
    \item Funding (industrial)
    \item Funding (governmental)
    \item Funding (academic)
\end{itemize}

\begin{figure}[t]
    \centering
    \includegraphics[width=\linewidth]{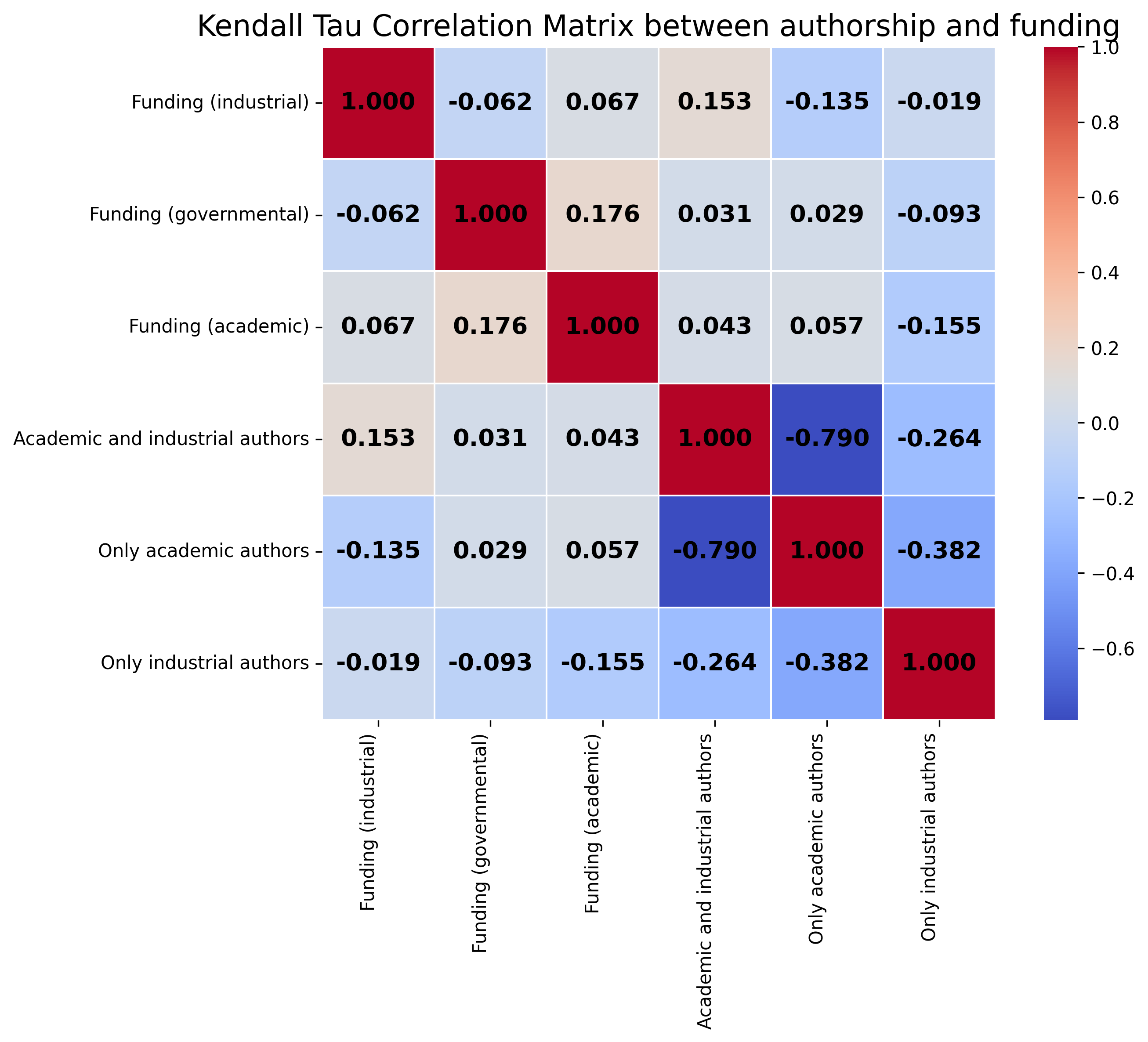}
    \caption{This figure displays the correlation between authors from industrial and academic backgrounds within various types of funding such as academic, industrial and governmental.}
    \label{fig:corr_funds}
\end{figure}

As seen in the Figure \ref{fig:corr_funds}, we see that there is no major correlation between funding and authorship. Thus it does not imply that if an industrial/academic author is present in the list of authors, there will necessarily be funding from the industry/academia/government. This finding is surprising actually as one would expect that a paper written by only academics would have academic/governmental funding and the same for industrial authors and industrial funding.

\begin{table}[ht]
\centering
\caption{Summary of Authors and Funding Categories}
\begin{tabular}{lr}
\toprule
\textbf{Category}                & \textbf{Sum} \\
\midrule
Academic and Industrial          & 78         \\
Only Academic authors            & 118        \\
Only Industrial authors          & 25         \\
Funding (industrial)             & 49         \\
Funding (governmental)           & 61         \\
Funding (academic)               & 35         \\
\bottomrule
\end{tabular}
\end{table}

\newpage

\section{Survey Dimensions, Methodologies, and Task Categories in Financial NLP Research}
\label{app:dimensions}

This appendix consolidates the key reference tables from our study, providing an overview of the dimensions examined (Table \ref{tab:original_dimensions}), the methodologies employed (Table \ref{tab:methodologies}), and the task categories identified in NLP research applied to the financial domain (Tables \ref{Tab:Finance Related} and \ref{Tab:Financial Contribution}).

\clearpage

\begin{table*}[t]

\centering
\footnotesize

\caption{This table outlines the dimensions examined in our analysis of NLP research focused on the financial domain. Dimensions that led to significant findings are discussed in the main paper, while  an analysis of authorship and funding is provided in Appendix \ref{app:authorship_funding}.}
\label{tab:original_dimensions}

\begin{tabular}{lp{0.73\textwidth}}
\toprule
\textbf{Exploration dimension} & \textbf{Description} \\
\midrule

Primary task & The main task addressed in the paper, such as volatility prediction \citep{wang-etal-2024-ama}. \\

Sub-Tasks & Additional tasks that support the primary objective, like using sentiment analysis to enhance stock price predictions \citep{jain-agrawal-2024-fb}. \\

Methodology & Techniques applied, ranging from traditional machine learning to deep learning and large language models. See Section \ref{sec:methodologies}. \\

Code and data availability & Whether the paper provides open-source code or data, and the quality of this accessibility (e.g., active links). \\

Contribution & What type of contribution does the paper have (Dataset, framework, model, evaluation, etc.)? \\

Comparability of research & What other research did the researchers compare their work to? \\

Data source & The types of data used, including financial reports, social media, news articles, or other financial documents. \\

Metrics & Evaluation criteria, including standard ML metrics (accuracy, precision) and finance-specific metrics (Sharpe ratio). \\

Data years & The time periods covered by the data, such as crisis years or more stable financial periods. \\

Authorship & Indicates if the paper was authored or co-authored by industry professionals, reflecting practical applications. \\

Funding source & The origin of the research funding (academic, industrial, or governmental) and its influence on the research focus. \\

\bottomrule
\end{tabular}

\end{table*}

\begin{table*}[t]

\caption{This table categorizes the key methodologies into nine groups.``PLMs'' stands for Pretrained Language Models, while ``LLMs'' stands for Large Language Models.}
\label{tab:methodologies}

\centering
\footnotesize

\begin{tabular}{lp{0.78\textwidth}}

\toprule

\textbf{Methodology} & \textbf{Description} \\

\midrule

Statistical NLP & Methods like TF-IDF, Bag-of-Words, and n-grams focused on extracting statistical patterns from financial text. \\

Embeddings & Techniques such as Word2Vec, GloVe, and custom embeddings designed to map financial terms into vector spaces, improving downstream tasks. \\

Conventional ML & Algorithms like Logistic Regression, Support Vector Machines (SVM), and Decision Trees, often used for classification and risk prediction. \\

Deep Learning (DL) & Custom neural network architectures tailored for specific financial tasks such as stock price forecasting. \\

PLMs & Encoder-only models (e.g., BERT) and Encoder-Decoder models (e.g., T5) are used for tasks like document summarization and information extraction. \\

LLMs & Models like GPT-3.5 and LLaMA-2 focus on text generation and understanding complex financial language. \\

Statistical Modeling & Involves correlation analysis, Granger causality, etc., to understand relationships in financial data. \\

Graphs & Uses graph structures to model interactions in financial systems. \\

Knowledge Graphs & Integrating structured knowledge within financial tasks to enhance model performance and reliability. \\

\bottomrule

\end{tabular}

\end{table*}

\clearpage

\begin{table*}[ht]
\centering
\scriptsize
\begin{tabular}{p{0.3\textwidth}p{0.65\textwidth}}

\toprule

\textbf{Primary Task} &\textbf{ Examples \&  Insights} \\

\midrule

\multicolumn{2}{c}{\textcolor[RGB]{150,25,255}{\textbf{Finance-Related (Categories IV)}}} \\

\midrule

Explainable AI (XAI) refers to methods that make machine learning outputs more interpretable and less of a "black box." & These papers show how small input changes can drastically affect model explanations \citep{sinha-etal-2021-perturbing} and highlight the instability of tools like LIME in sensitive applications \citep{burger-etal-2023-explanations}. In finance, where decisions must be traceable and justifiable, clear and reliable explanation methods are essential for building trust in tasks like fraud detection and risk assessment. \\

Privacy-preserving methods protect sensitive data while still allowing model training and analysis. & Examples include homomorphic encryption for text similarity \citep{kim-etal-2022-toward}, federated learning for distributed training without raw data exchange \citep{zhao-etal-2024-source}, and domain adaptation that shares only model parameters \citep{xiao-etal-2024-federated}. These methods are especially important in finance, where institutions must work with private data while staying compliant and avoiding breaches. \\

Numerical reasoning involves working with numbers and structured data to solve problems or develop algorithms. & This includes solving math word problems using declarative knowledge \citep{roy-roth-2018-mapping}, learning numeracy through number embeddings \citep{duan-etal-2021-learning-numeracy}, and pretraining models for verifying tabular claims \citep{akhtar-etal-2023-exploring}. In finance, good numerical reasoning supports tasks like market prediction and risk evaluation by improving how models interpret and reason about numbers. \\

\midrule

\multicolumn{2}{c}{\textcolor[RGB]{50,255,50}{\textbf{Financial Applications (Category III)}}} \\

\midrule

Sentiment analysis helps assess market mood and forecast financial movements using sources like news, social media, and reports. & Studies analyze opinions in reports and news \citep{rodriguez-inserte-etal-2023-large}, investor sentiment on social media \citep{guo-etal-2023-predict}, and tone in financial texts \citep{choe-etal-2023-exploring}. As markets become more influenced by public opinion, sentiment analysis plays a growing role in trading and analysis. \\

Information extraction (IE) and relation extraction focus on identifying entities and linking them to events or other entities. & Examples include extracting financial events from documents \citep{zheng-etal-2019-doc2edag}, signals in reports \citep{huang-etal-2023-fish}, and patterns in social media \citep{conforti-etal-2022-incorporating}. These tools help connect companies, individuals, and financial instruments to real-world developments \citep{liou-etal-2021-dynamic}, supporting real-time analysis and prediction. \\

Question Answering (QA) enables systems to find specific information in financial documents, reports, and databases. & QA models support quick access to relevant facts \citep{mavi-etal-2023-retrieval}, and recent work focuses on calibrating large language models (LLMs) for finance-specific tasks \citep{zhao-etal-2024-optimizing,theuma-shareghi-2024-equipping,addlesee-2024-grounding}. This area emphasizes not only building QA tools but also improving their accuracy for use in high-stakes financial contexts. \\

\bottomrule

\end{tabular}
\caption{Overview of key primary tasks in \textcolor[RGB]{150,25,255}{\textbf{Categories IV (Finance-Related)}} above the mid line and \textcolor[RGB]{50,255,50}{\textbf{Category III (Financial Applications)}} below. Each entry summarizes the task's role and relevance in NLP research applied to finance, with representative examples and practical insights.}
\label{Tab:Finance Related}
\end{table*}

\begin{table*}[ht]
\centering
\scriptsize

\begin{tabular}{p{0.30\textwidth}p{0.65\textwidth}}

\toprule

\textbf{Primary Task} & \textbf{Examples \& Insights} \\

\midrule

\multicolumn{2}{c}{\textcolor{blue}{\textbf{Financial Resources (Category II)}}} \\

\midrule

Dataset and Resource Construction involves building structured, labeled financial datasets for model development and evaluation. & These papers focus on creating large-scale annotated resources \citep{chen-etal-2020-issues}, including datasets like Tab-CQA \citep{liu-etal-2023-tab} and ConvFinQA \citep{chen-etal-2022-convfinqa} designed for reasoning over tables and multi-step numerical queries. Other works assemble corpora from financial reports \citep{shah-etal-2022-flue}, company filings \citep{zmandar-etal-2022-cofif}, news \citep{tang-etal-2023-finentity}, and government documents \citep{shah-etal-2023-trillion}, enabling downstream tasks like entity linking, numerical reasoning, or forecasting. \\

Fraud detection leverages domain-specific data and models to identify deceptive financial behavior across platforms. & \citet{erben-waldis-2024-scamspot} identifies financial scams on Instagram using a fine-tuned BERT model deployed via browser extension and REST API. Another approach detects identity fraud through interactive dialogue, employing knowledge graphs and reinforcement learning to expose inconsistencies in claimed personal data \citep{wang-etal-2019-real}. These systems highlight how NLP architectures can protect users from financial manipulation. \\

\midrule

\multicolumn{2}{c}{\textcolor{red}{\textbf{Financial Forecast (Category I)}}} \\

\midrule

Stock Price and Volatility Prediction aims to forecast stock movements or market instability using text data. & These models use historical data \citep{sawhney-etal-2021-fast}, news \citep{ahbali-etal-2022-identifying}, and event-driven signals \citep{sawhney-etal-2020-deep, wu-2020-event} to predict stock trends. Volatility prediction further explores market sensitivity by analyzing unstructured text like press releases or financial articles \citep{qin-yang-2019-say}, helping traders anticipate fluctuations. \\

Risk Prediction involves identifying the likelihood and impact of adverse financial events, such as defaults or market disruptions. & These models analyze unstructured data like earnings call transcripts \citep{sang-bao-2022-dialoguegat}, regulatory filings, and legal documents to detect early signals of financial risk \citep{li-etal-2023-stinmatch}. Applications include credit risk estimation, fraud detection, and systemic risk monitoring \citep{zhang-etal-2024-finbpm}. \\

\bottomrule

\end{tabular}
\caption{Descriptions of major primary tasks in \textcolor{blue}{\textbf{Category II (Financial Resources)}} above the line and \textcolor{red}{\textbf{Category I (Financial Forecast)}} below. Each entry summarizes the task’s focus and contribution to NLP research applied to finance, with representative examples.}
\label{Tab:Financial Contribution}
\end{table*}

\clearpage

\section{Conference Proceedings}
\label{app:conferences}

Table \ref{tab:conf_categories} categorizes key conferences and workshops according to the four primary task categories identified in our survey.

\section{Papers per year}
\label{app:papers_year}

Table \ref{tab:papers_per_year} reports the number of papers per year from 1975 to 2024.

\section{Glossary}
\label{app:Glossary}

Table \ref{tab:metrics_definitions} provides definitions to the key financial metrics mentioned in this paper.

\begin{table*}[ht]
    \centering
    \begin{tabular}{p{0.4\textwidth}p{0.6\textwidth}}
        \toprule
        \textbf{Category} & \textbf{Conferences and Workshops} \\
        \midrule
        \textcolor{red}{\textbf{Financial Forecast (Category I)}} &
        ACL, Australasian Language Technology Association Workshop, CCL, EACL, EcoNLP Workshop, 
        EMNLP, EMNLP (Industry), Financial Technology and Natural Language Processing (FinNLP) Workshop, 
        ICCL, LREC-COLING, NAACL, NAACL (Industry), SRW \\
        \midrule
        \textcolor{blue}{\textbf{Financial Resources (Category II)}} &
        ACL, ACL (Industry), CoNLL, EACL, EcoNLP Workshop, EMNLP, EMNLP-IJCNLP, 
        e-Commerce and NLP Workshop, Financial Narrative Processing and MultiLing Financial Summarisation Workshop, 
        Financial Technology and Natural Language Processing (FinNLP) Workshop, GWC, ICON, ICCL, LREC, LREC-COLING, NAACL, NAACL (Industry), NLP4PI Workshop, Pattern-based Approaches to NLP in the Age of Deep Learning Workshop, 
        SIGHUM Workshop \\
        \midrule
        \textcolor[RGB]{50,255,50}{\textbf{Financial Applications (Category III)}} &
        ACL, ATALA, Bridging Human–Computer Interaction and Natural Language Processing Workshop, 
        CCL, CoLM, Computational Approaches to Subjectivity, Sentiment and Social Media Analysis Workshop, 
        DeeLIO Workshop, EACL, EcoNLP Workshop, EMNLP, EMNLP (Industry), EMNLP-IJCNLP, 
        Financial Narrative Processing and MultiLing Financial Summarisation Workshop, 
        Financial Technology and Natural Language Processing (FinNLP) Workshop, ICON, ICCL, INLG, 
        LREC, LREC-COLING, NAACL, NAACL (Industry), Natural Legal Language Processing Workshop, 
        News Media Content Analysis and Automated Report Generation Workshop, Pattern-based Approaches to NLP in the Age of Deep Learning Workshop, 
        Safety4ConvAI Workshop, Structured Prediction for Natural Language Processing Workshop, TextGraphs \\
        \midrule
        \textbf{\textcolor[RGB]{150,25,255}{\textbf{Finance Related (Category IV)}}} &
        ACL, ACL-IJCNLP, BlackboxNLP Workshop, EMNLP, 
        Financial Technology and Natural Language Processing (FinNLP) Workshop, 
        LREC-COLING, SRW, TrustNLP Workshop \\
        \bottomrule
    \end{tabular}
    \caption{Conference and Workshop Categorization}
    \label{tab:conf_categories}
\end{table*}

\begin{table*}[ht]
\centering
\resizebox{\textwidth}{!}{%
\begin{tabular}{lccccccccccccccccccc}
\toprule
Year & 2024 & 2023 & 2022 & 2021 & 2020 & 2019 & 2018 & 2017 & 2016 & 2015 & 2014 & 2012 & 2010 & 2008 & 2006 & 2002 & 2001 & 1976 & 1975 \\
\midrule
Number of Papers & 127 & 53 & 90 & 32 & 57 & 15 & 14 & 21 & 6 & 1 & 5 & 1 & 3 & 3 & 1 & 1 & 1 & 1 & 1 \\
\bottomrule
\end{tabular}
}
\caption{Number of papers satisfying our filtration criterion per year.}
\label{tab:papers_per_year}
\end{table*}

\begin{table*}[ht]
\centering
\begin{tabular}{@{}p{3.5cm}p{10cm}@{}}
\toprule
\textbf{Metric} & \textbf{Definition} \\
\midrule
Sharpe Ratio & Measures how much return an investment gives for each unit of risk, compared to a risk-free asset~\citep{Investopedia_SharpeRatio}. \\
Maximum Drawdown & The biggest drop from a peak to a low point in a portfolio's value before it recovers. \\
Cumulative Return & The total profit or loss from an investment over time. \\
\bottomrule
\end{tabular}
\caption{Key Definitions of financial terms used in our paper.}
\label{tab:metrics_definitions}
\end{table*}
\section{Data Sources}
\label{app:DataSources}
\begin{table*}[h!]
\centering
\renewcommand{\arraystretch}{1.2}
\resizebox{\linewidth}{!}{
\begin{tabular}{p{0.14\textwidth} p{0.28\textwidth} p{0.30\textwidth} p{0.12\textwidth} p{0.14\textwidth}}
\hline
\textbf{Document type} & \textbf{Sources} & \textbf{Accessibility (incl.\ access method)} & \textbf{Modalities} & \textbf{Update cadence} \\
\hline
Regulatory filings &
SEC EDGAR; other registries: SEDAR (Canada), EDINET (Japan), Companies House (UK) &
Public via web UI; free EDGAR APIs and RSS \citep{SEC_EDGAR_APIs}; other registries public but less commonly used &
Text/PDF &
Event-driven; periodic (annual/quarterly) \\
\hline
Earnings calls (transcripts, slides, audio) &
Seeking Alpha \citep{seekingalpha}; The Motley Fool \citep{motleyfool}; FactSet \citep{factset}; Thomson Reuters \citep{thomsonreuters}; S\&P Global \citep{spglobal}; company IR sites &
Mixed: many providers behind paywalls; IR sites often public; access via web UI and provider APIs (where offered) &
Text transcripts; audio &
Quarterly around earnings; event-driven (scheduled call dates) \\
\hline
Financial news &
Reuters, Bloomberg, Dow Jones, CNBC, FT, WSJ; free aggregators like Yahoo Finance \citep{yahoofinance} and Google News \citep{googlenews} &
Mixed: premium (e.g., Bloomberg) and free sources, including static datasets \citep{reuters21578,fnspid2024}; access via web UIs and vendor APIs (e.g., Bloomberg API \citep{bloombergapi}) &
Text (articles; headlines) &
Real-time / continuous \\
\hline
Social media &
X (Twitter); Reddit; sentiment datasets for X \citep{sentiment140, yash_chaudhary_2020,xu-cohen-2018-stock} and Reddit \citep{redditdataset, wang2024socialnetworkdatasetsreddit} &
X API with limited free and paid tiers \citep{xapi}; StockTwits free/commercial API  \citep{stocktwits}; public web UIs &
Short-text posts; sentiment tags (StockTwits Bullish/Bearish) &
Real-time \\
\hline
Macroeconomic \& monetary communications &
U.S. Federal Reserve (FOMC) statements and transcripts \citep{fomcsource} &
Public via official web portals; archives available &
Text (policy statements, minutes, press releases) &
Scheduled (policy meetings, release calendars) \\
\hline
Economic indicators &
BEA \citep{bea}, FRED \citep{FRED}, the World Bank \citep{worldbank} &
Public via official portals and APIs &
Numeric time series + accompanying text commentary &
Scheduled (monthly / quarterly) \\
\hline
Analyst reports &
Bloomberg Terminal \citep{bloombergterminal}; S\&P Capital IQ \citep{capitaliq} &
Paid/proprietary subscription platforms; enterprise access/APIs; occasional public excerpts in news/Seeking Alpha &
Text/PDF (reports, notes) &
Ongoing; event-driven (coverage initiations, earnings, sector updates) \\
\hline
\end{tabular}
}
\caption{Financial text/data sources by document type with sources, accessibility (and access method), modalities, and update cadence.}
\end{table*}

\paragraph{Regulatory Filings and Earnings Calls} U.S. public companies file a range of documents such as 10‑K, 10‑Q, 8‑K, credit agreements, proxy statements, S‑1, S‑3, and others are accessible via the SEC’s EDGAR platform, through free API calls and RSS feeds \citep{SEC_EDGAR_APIs}. While similar registries exist (e.g., Canada’s SEDAR, Japan’s EDINET, UK’s Companies House), they are rarely used, reinforcing a U.S.-centric and English-language bias (Section \ref{subsec:data_biases}). Earnings calls, where firms discuss quarterly performance with investors, are another rich data source \cite{mukherjee-etal-2022-ectsum, koval-etal-2023-forecasting, pardawala2025subjectiveqameasuringsubjectivityearnings}. Seeking Alpha transcribes ~4,500 calls quarterly \citep{seekingalpha}; other providers include The Motley Fool \citep{motleyfool}, FactSet \citep{factset}, Thomson Reuters \citep{thomsonreuters}, and S\&P Global \citep{spglobal}, though often behind paywalls. Companies also post transcripts/slides on IR websites. Public datasets include transcriptions and audio, such as S\&P 500 calls from 2017 \citep{qin-yang-2019-say}  and 1,213 companies' data covering 2015-2018 \citep{10.1145/3340531.3412879}.

\paragraph{Financial News and Social Media} Real-time news from sources like Reuters, Bloomberg, Dow Jones, CNBC, FT, and WSJ is vital for forecasting. The Reuters-21578 dataset \citep{reuters21578}, with 10,369 articles from 1987, remains a benchmark in text analysis. Larger archives such as FNSPID \citep{fnspid2024} link 15.7 million articles (1999-2023) to S\&P 500 stocks. News is accessible via APIs (e.g., Bloomberg \citep{bloombergapi}) and free sources like Yahoo Finance \citep{yahoofinance} and Google News \citep{googlenews}. On social media, “FinTwit” and platforms like X and StockTwits are used for sentiment analysis. Key datasets include Sentiment140 \citep{sentiment140, yash_chaudhary_2020}, StockNet \citep{xu-cohen-2018-stock}, and Reddit datasets \citep{redditdataset, wang2024socialnetworkdatasetsreddit}. X’s API offers limited free and full historical access under paid tiers \citep{xapi}. StockTwits, with Bullish/Bearish sentiment tagging, offers free and commercial API access \citep{stocktwits} and is often used in forecasting studies.

\paragraph{Macroeconomic and Monetary Data} Central bank communications (policy statements, minutes, and press releases) offer critical macroeconomic guidance \cite{ahrens-mcmahon-2021-extracting, peskoff-etal-2023-gpt-fedspeak, menzio-paris-fersini-2024-unveiling}. The U.S. Federal Reserve’s FOMC releases detailed statements and transcripts on interest rates, inflation, and economic outlook \citep{fomcsource}, with archives available on the Fed’s website. Despite their market relevance, such texts remain underused in NLP-based forecasting. \citet{shah-etal-2023-trillion} addresses this by introducing an annotated dataset of FOMC communications and proposing a sentence-level classification task (hawkish, dovish, neutral) using fine-tuned transformers. Economic indicators like GDP, CPI, and unemployment figures -- issued by the BEA \citep{bea}, FRED \citep{FRED}, Eurostat \citep{eurostat}, IMF \citep{imf}, OECD \citep{oecd}, and the World Bank \citep{worldbank} -- are mostly numerical but often include press commentary, all freely available via official portals.

\paragraph{Analyst Reports} These are proprietary equity research reports with buy/sell calls and sector outlooks. Data providers such as Thomson Reuters, FactSet, Bloomberg, and S\&P I/B/E/S aggregate these reports, typically through subscription services. All-in-one platforms like the Bloomberg Terminal \citep{bloombergterminal} and S\&P Capital IQ \citep{capitaliq} provide access to equity research, earnings estimates, and corporate event data as part of their premium offerings. There is no comprehensive free source; however, excerpts or key takeaways occasionally surface in news articles or on platforms like Seeking Alpha.

\end{document}